\documentclass[11pt]{aime}
\usepackage{booktabs}
\usepackage{array}
\usepackage{xcolor}

\Crefname{appendix}{Appendix}{Appendices}
\crefname{appendix}{appendix}{appendices}
\newcommand{\appfig}[1]{Appendix~Fig.~\ref{#1}}

\title{Can phenotypic activity be predicted without experimental readouts?}

\author[1]{Télio Cropsal}
\author[1,2,*]{Rocío Mercado}

\affiliation[1]{AI Laboratory for Molecular Engineering (AIME), Department of Computer Science and Engineering, Chalmers University of Technology \& University of Gothenburg, Gothenburg, Sweden}
\affiliation[2]{Science for Life Laboratory (SciLifeLab), Gothenburg, Sweden}

\contribution[*]{Corresponding author}  

\correspondence{\email{rocio.mercado@chalmers.se}}
\codeavailability{\url{https://github.com/cctelio/predicting-phenotypic-activity}}
\date{\today}

\abstract{%
Molecular encoders contrastively pretrained on paired molecule-morphology data, such as CLOOME and CellCLIP, have been proposed as cheap surrogates for phenotypic prediction, avoiding the need to run a Cell Painting assay. We evaluate this idea under a protocol designed to control for two confounds that can inflate apparent performance: leakage across an encoder's own pretraining boundary, and the correlation between phenotypic activity and simple cytotoxicity. Testing six representations, including a non-pretrained MLP control matching CLOOME's input and layer count, on two Cell Painting screens of different scale and origin, we find that once these confounds are controlled for, CLOOME and CellCLIP show no clear advantage over plain physicochemical descriptors, also under stricter activity definitions and with non-linear probes, and that toxicity is generally easier to predict than genuine phenotypic activity across representations. The protocol is not tied to these two models, and we suggest that leakage-aware, confound-controlled evaluation should be standard practice before phenotype-pretrained encoders are trusted as surrogates for predictive phenotypic drug discovery.
}

\begin{document}
\maketitle


\section{Introduction}
\label{sec:intro}

Cell Painting and related high-content phenotypic assays report how a compound perturbs cell biology across hundreds of morphological features, but most of that effort is spent on compounds that are phenotypically inactive \cite{seal2025cell}. Being able to accurately predict activity from structure, before the plate is run, would let a screening campaign spend its budget more efficiently by prioritizing phenotypically active compounds \cite{marin2023cell, fredin2024cell}. Predicting this activity, however, is a challenging task. \citet{moshkov2023predicting} show that morphology and chemical structure carry weak and largely non-overlapping predictive signal, with morphology or structure predicting $\le$28 of 270 assays at AUROC $>0.9$. Contrastive molecule-morphology pretraining is presented as a way to bridge these complementary modalities: encoders such as CLOOME \cite{sanchez2023cloome} and CellCLIP \cite{lu2026cellclip} are trained to align structures with the morphological profiles they induce in assays, and have been proposed as surrogates that inherit phenotypic information without needing to run an assay \cite{harrison2026biologically}.

Two issues complicate a fair evaluation of such surrogates. First, phenotypic activity does not distinguish a specific, mechanistically interesting response from nonspecific cytotoxicity: a compound that simply reduces cell number produces a large and easily detected morphological deviation from control, so a nontrivial share of active calls in a Cell Painting screen can be driven mainly by cell death rather than a target-relevant phenotype. Compound physicochemical properties, lipophilicity in particular, are well known to correlate with nonspecific cytotoxicity \cite{lu2012contribution}, so a surrogate that appears to predict phenotypic activity well may in fact be learning to predict toxicity, a substantially easier and less interesting target. \citet{seal2026counting} recently quantified this: cell count alone predicts a substantial share of Cell Painting bioactivity benchmarks, and CLOOME embeddings score comparably to a cell-count baseline and below cell count combined with molecular weight and LogP. Second, a molecular encoder's apparent ability to predict phenotype can come from several different places, and a single train/test split cannot tell them apart: an encoder such as CLOOME \cite{sanchez2023cloome} or CellCLIP \cite{lu2026cellclip} may be recalling specific compounds it saw during its own contrastive pretraining rather than generalizing to new ones \cite{fradkin2024molecules}, it may be picking up on a plate or batch-specific technical artifacts \cite{moshkov2024learning} that happen to correlate with the label rather than chemistry, or it may simply never be tested on data different enough from its training distribution for either failure to show up. Telling real generalization from these artifacts apart requires testing across several different notions of new at once: new compounds, new experimental batches, and, ideally, an entirely different laboratory.

In this work we evaluate structure-based and phenotype-pretrained molecular encoders as surrogates for phenotypic activity under a design intended to control for both issues directly. We construct two labels, phenotypic activity and toxicity, using the same permutation-testing methodology so that we can isolate the harder, more interesting target of non-cytotoxicity-driven activity. We evaluate every encoder, including two contrastively pretrained molecule-morphology encoders, on chemical and batch-generalization splits as well as, where available, the encoder's own official pretraining-respecting held-out split, so that any reported performance can be traced to a specific, well-defined notion of generalization rather than a single, potentially leakage-confounded number. We repeat this evaluation across two Cell Painting resources of very different scale, chemical diversity, and origin: a curated, single-laboratory drug-repurposing screen and a much larger, chemically diverse, multi-laboratory discovery screen, so that any conclusion is not an artifact of one screen's particular chemical or experimental composition.

We make three key contributions:
\begin{itemize}
    \item A confound-controlled labeling scheme that statistically separates phenotypic activity
    from toxicity using the same validated permutation-test framework, yielding disentangled
    ``active, not toxic'' labels.
    \item A leakage-aware evaluation protocol for phenotype-pretrained molecular encoders,
    comparing chemical and batch-generalization splits against the encoder's own official
    pretraining-respecting split, under a statistically rigorous repeated cross-validation and
    Tukey's honestly significant difference comparison protocol appropriate for the size of the
    datasets involved.
    \item Cross-dataset validation spanning two Cell Painting resources with very different scale,
    chemical diversity, and laboratory origin, testing whether conclusions about the value of
    phenotype-pretrained molecular encoders generalize beyond a single screen.
\end{itemize}


\section{Methods}
\label{sec:methods}

\begin{figure}[H]
    \centering
    \includegraphics[width=0.99\linewidth]{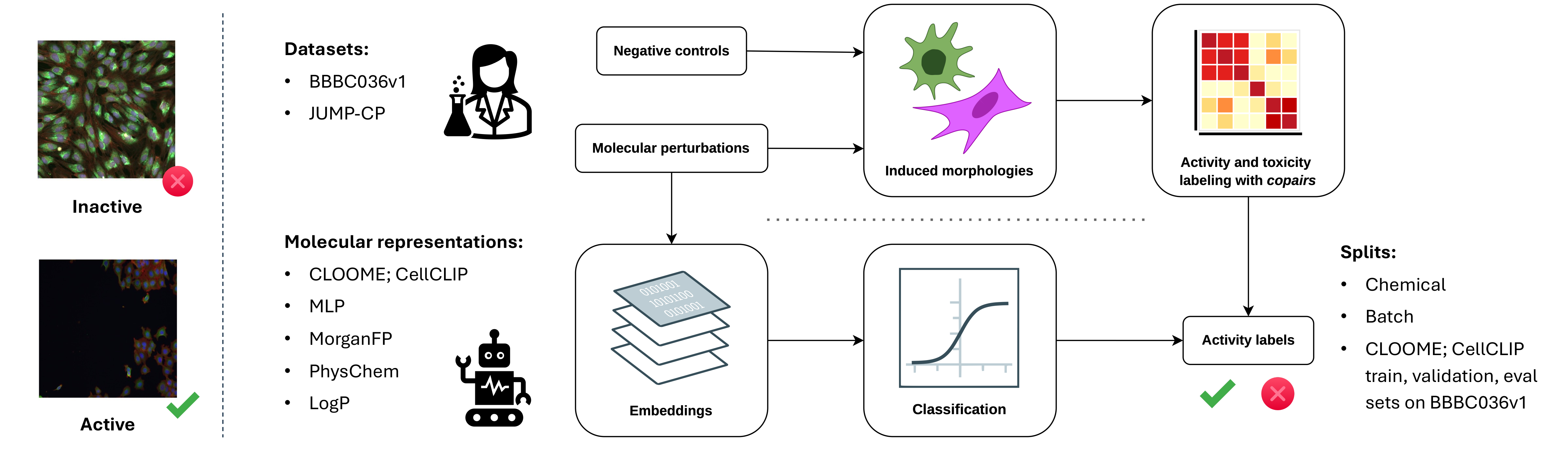}
    \caption{%
        A molecular perturbation can be used two ways towards the same classification task. Experimentally (top), the compound is applied to cells and its effect is captured as a Cell Painting morphological profile; \texttt{copairs} \cite{kalinin2025versatile} permutation testing on that profile separates truly active compounds from inactive ones, and the same test applied to cell count alone flags toxicity, giving the ground-truth label. Computationally (bottom), the same classification is predicted directly from a molecular representation of the compound, physicochemical descriptors, a fingerprint, or a phenotype-pretrained embedding like CLOOME or CellCLIP.
    }
    \label{fig:method-overview}
\end{figure}

\Cref{fig:method-overview} summarizes the workflow presented in this work.

\subsection{Data}
\label{sec:data}

\paragraph{} \textbf{Label construction} Both activity and toxicity are tested with \texttt{copairs} \cite{kalinin2025versatile}: a compound's replicate wells are scored by average precision against a permutation-derived null ($10{,}000$ permutations), using up to $190$ sampled negative control wells per plate as the negative reference and Benjamini-Hochberg-corrected $p$-values ($q$) across compounds, with compounds called at $q<0.05$. Activity is tested on the morphological profile with cosine distance; following \citet{seal2026counting}, toxicity is tested on cell count alone, standardized per plate against negative control, with Euclidean distance instead of cosine. We use this cell-count-based toxicity proxy because it is measured in the same wells as the morphological profiles and available for every compound without an external cytotoxicity assay; we refer to it as toxicity for brevity. The toxicity test is restricted to compounds already called active, since toxicity in the absence of activity is essentially never observed and the restriction keeps the well level data volume tractable at JUMP-CP's scale. Every compound is thus assigned to one of three categories evaluated throughout: \textbf{inactive}, \textbf{active and toxic}, or \textbf{active and not toxic}. The last one is our main target. To test how much the results depend on the activity threshold, we also call activity at $q<0.01$ and $q<0.001$, keeping toxicity at $q<0.05$ (\appfig{fig:activity}).

\paragraph{} \textbf{BBBC036v1} \cite{bray2017dataset} is a $\sim$30,000-compound screen of human U2OS cells with CellProfiler-derived morphological profiles (1,449 features per well, DMSO as negative control). We use the well-level profiles distributed with CPMolGAN \cite{marin2023cell} and the official compound-level train/validation/test split leveraged by CLOOME and CellCLIP (10,683 compounds). Well-level profiles are corrected with typical variation normalization (TVN) \cite{ando2017improving} against per-plate DMSO controls, with the cell-count feature explicitly excluded from the feature set before whitening. Morphology is aggregated to the compound level as the median TVN-corrected profile
across a compound's replicate wells and both activity and toxicity are computed. After joining to compounds with an available
morphological profile, the working population is 10,680 compounds, of which $91.3\,\%$ are called active; of those, $6.3\,\%$ are additionally called toxic.

\paragraph{} \textbf{JUMP-CP} \cite{chandrasekaran2023jump} is a much larger, chemically diverse, multi-laboratory compound screen also performed in human U2OS cells. Recomputing multivariate phenotypic activity ourselves at JUMP-CP's scale would require reprocessing the full well-level profile for on the order of $100$k compounds; we instead reuse the corrected $p$-values published by \textit{jump\_rr}, computed with the same style of permutation-testing methodology described above. Toxicity, by contrast, is computed directly by our own procedure, restricted to the resulting active
compounds. SMILES and compound metadata are taken from the official JUMP-CP metadata release. We exclude any compound whose first-14-character InChIKey matches a compound in BBBC036v1's official population (1,540 of 115,779 compounds), so that the BBBC036v1-pretrained encoders are evaluated on JUMP-CP on compounds they could not have seen during their own pretraining. Morphology is taken from the JUMP consortium's own Harmony-batch-corrected, feature-selected profile product (722 features), median-aggregated per compound across replicate wells; this profile carries no cell-count feature, so toxicity is instead computed from the JUMP consortium's separate interpretable profiles. Under this procedure, $46.2\,\%$ of active JUMP-CP compounds are additionally called toxic, a substantially higher rate than BBBC036v1's, consistent with JUMP-CP being a chemically broader, less pre-filtered discovery screen than BBBC036v1's curated library.

Of JUMP-CP's twelve data-generating sources, ten have compound-level activity $p$-values with \textit{jump\_rr} (the other two run only CRISPR and ORF genetic perturbations). One of these ten, \textit{source\_7}, is a curated library of compounds, assayed at $0.625\,\mu$M versus $10\,\mu$M for the other sources \cite{munoz2026jump}; its activity rate reflects this ($23.0\,\%$ of its 5,204 compounds, roughly twice the typical rate of the other sources). Rather than let this different compound population and concentration confound the batch-generalization split, we exclude it and evaluate it separately. The remaining population is dominated by inactive compounds ($10.1\,\%$ active), so we construct the working population as every active compound (10,975) plus a matched, source-stratified random sample of an equal number of inactive compounds, giving $n=21{,}950$, approximately class-balanced by construction and sampled proportionally across the remaining 9 sources.

\paragraph{} \textbf{Activity in the two screens} The two screens differ sharply in how many compounds are called active (\appfig{fig:activity}). In BBBC036v1, $91.3\,\%$ of compounds are active, most with a high mAP, the \texttt{copairs} effect size (median $0.70$), whereas only $10.7\,\%$ of JUMP-CP's full compound screen is active (median mAP $0.17$; mAP values are not directly comparable between screens with different numbers of replicates and controls). At the stricter thresholds, the active fraction drops to $79.6\,\%$ and $50.0\,\%$ in BBBC036v1, against $7.5\,\%$ and $5.2\,\%$ in JUMP-CP. Together with the toxicity rates above, BBBC036v1 thus has many actives that are rarely toxic, and JUMP-CP few actives, about half of them toxic.

\subsection{Molecular representations}
\label{sec:model}

Every prediction target is evaluated identically across all representations below, each standardized (zero mean, unit variance) before fitting, with statistics computed on the training fold only. 

\paragraph{} \textbf{CLOOME} is a 4-layer multilayer perceptron mapping chiral Morgan fingerprints (radius 3, 1024 bits) to a 512-dimensional embedding, trained via contrastive learning on BBBC036v1 molecule-morphology pairs. We use the publicly released bioactivity checkpoint (\texttt{anasanchezf/cloome}).

\paragraph{} \textbf{CellCLIP} is a BERT-base-cased text encoder with a linear projection to 512
dimensions, trained via contrastive learning on paired Cell Painting images and text. As with CLOOME, we compute its embeddings via a forward pass through the model with the publicly released
checkpoint (\texttt{suinleelab/CellCLIP} on Hugging Face). Prompts are templated as \texttt{"SMILES: \{smiles\}"} only, no compound name and no
cell-type conditioning, so CellCLIP is given exactly the same input information (structure alone)
as every structure-only baseline evaluated.

\paragraph{} \textbf{Structure-only and physicochemical baselines} require no morphology-derived training signal, so any gap between them and CLOOME/CellCLIP that survives the leakage-aware evaluation is attributable to what the contrastive pretraining actually added, not merely to using richer chemical information.
\begin{itemize}
    \item \textbf{MLP}: a 4-layer multilayer perceptron over the same chiral Morgan fingerprint input as CLOOME, matching its input dimensionality and layer count (though not its exact per-layer widths; see \Cref{app:model-details}), but trained directly and only on our supervised labels, with no contrastive pretraining. It serves as a no-pretraining control for CLOOME rather than as a probe.
    \item \textbf{MorganFP}: radius-2, 2048-bit Morgan fingerprint, used directly as a linear-probe
    input and, via its pairwise Tanimoto distance, to construct the chemical-generalization split.
    \item \textbf{PhysChem}: the full RDKit descriptor set (217 descriptors).
    \item \textbf{LogP}: a single Crippen logP value \cite{wildman1999prediction}, kept as its own representation to isolate the physicochemical property most classically linked to nonspecific bioactivity and cytotoxicity. Starting from LogP, we also refit the logistic-regression probe on a nested set of interpretable descriptors, adding molecular weight and heavy-atom count (size), then TPSA and hydrogen-bond donors and acceptors (polarity), then the number of basic amines, to see how much of PhysChem's signal these few properties recover (\appfig{fig:descriptors}).
\end{itemize}

\subsection{Evaluation protocol}
\label{sec:metrics}

\paragraph{} Each representation is evaluated under up to three different splits:
\begin{itemize}
\item \textbf{Chemical}: Butina clustering \cite{butina1999unsupervised} on the pairwise Tanimoto distance matrix of Morgan fingerprints (distance cutoff $0.35$, chosen so clusters are not so small that folds become
unbalanced), with whole clusters, not individual compounds, allocated to cross-validation folds via a size-balancing assignment, so that structurally similar compounds are largely kept on the same side of the split.
\item \textbf{Batch}: analogous grouped cross-validation over an experimental-batch identifier, assigned to each compound as the plate (BBBC036v1) or data-generating source (JUMP-CP) where it has the most replicate wells, since individual compounds are replicated across several plates, and on JUMP-CP often across several sources. This is one simple rule applied identically to both datasets, though a compound's other replicate wells may still have contributed to its aggregated profile. For JUMP-CP, since the batch identifier is the data-generating source, this tests generalization to an entirely different laboratory, not just a different plate within the same one. JUMP-CP's sources also differ in compound composition, experimental conditions, and activity rate, so this split tests generalization across all of these factors together rather than across laboratories alone. For JUMP-CP's bioactive-library sub-population (\Cref{sec:data}), which has only one source, the same rule is applied at the plate level instead, since source is no longer a meaningful grouping there.
\item \textbf{Official}: for BBBC036v1 only, the compound-level train/validation/test partition published alongside CLOOME and CellCLIP, which we use as provided. To compare it with our own splits, we report for each split the similarity of test compounds to the training set, label prevalence, and median physicochemical properties (\Cref{tab:splits}).
\end{itemize}

\begin{table}[H]
\centering
\footnotesize
\setlength{\tabcolsep}{4pt}
\caption{Train and test sets of the three BBBC036v1 splits. NN Tanimoto: similarity of each test compound to its closest training compound. Label percentages are over all compounds; labels and properties are given as train\,/\,test. Chemical and batch: mean over the 25 cross-validation folds.}
\label{tab:splits}
\begin{tabular}{@{}lccccccc@{}}
\toprule
\textbf{Split} & \textbf{Train\,/\,test} & \textbf{NN Tanimoto} & \textbf{NN $\geq 0.7$} & \textbf{Active} & \textbf{Active \& toxic} & \textbf{LogP} & \textbf{Mol. weight} \\
 & & \textbf{(median)} & \textbf{(\% of test)} & \textbf{(\%)} & \textbf{(\%)} & \textbf{(median)} & \textbf{(median)} \\
\midrule
Official & 7{,}479\,/\,2{,}131 & 0.49 & 15.9 & 91.5\,/\,90.4 & 5.7\,/\,5.3 & 3.31\,/\,3.36 & 357\,/\,351 \\
Chemical & 8{,}544\,/\,2{,}136 & 0.48 & 4.4  & 91.3\,/\,91.3 & 5.8\,/\,5.8 & 3.32\,/\,3.31 & 356\,/\,357 \\
Batch    & 8{,}544\,/\,2{,}136 & 0.51 & 16.9 & 91.3\,/\,91.3 & 5.8\,/\,5.8 & 3.32\,/\,3.31 & 356\,/\,357 \\
\bottomrule
\end{tabular}
\end{table}

\paragraph{} \textbf{Linear and non-linear probes} Each fixed representation is fed to an $\ell_2$-regularized logistic-regression probe, our main model: if phenotype-aware pretraining is useful, it should make phenotypic properties more linearly accessible than the raw molecular inputs do, and the same probe is applied to every representation, baselines included. To check that a linear probe does not understate the pretrained embeddings, CLOOME, CellCLIP, and PhysChem are also fed to a non-linear MLP probe: the MLP control's network, with its input layer matched to the dimension of the representation (\appfig{fig:probes}). The MLP control itself is trained directly on our labels from its Morgan fingerprint input (\Cref{app:model-details}).

\paragraph{} \textbf{Statistical comparison} For the chemical and batch splits, we use 5-times-repeated, 5-fold grouped cross validation, giving 25 performance samples per representation/target/split combination, following recommended practice for statistically powered method comparison at this data scale \cite{ash2025practically}. The same fold assignment is used across every representation within a repeat, so that representations are compared on identical held-out compounds. Logistic-regression probe regularization is selected per fold via an inner training/validation split; the MLP control is trained with dropout, weight decay, and early stopping on an analogous inner split. Representations are compared pairwise via repeated-measures analysis of variance followed by Tukey's honestly significant difference test, using the within-subject error term from the repeated-measures model rather than a naive independent-samples pooled variance, since every representation is evaluated on identical fold partitions; we report adjusted $p$-values and Cohen's $d$ effect sizes for every pairwise comparison. We check the repeated-measures ANOVA's sphericity assumption with Mauchly's test for every comparison and apply a Greenhouse-Geisser correction to the error degrees of freedom whenever it is rejected, which is the case for most comparisons in this study. For the official split, which by construction offers only a single train/validation/test partition, we approximate the same repeated-measures comparison by bootstrap-resampling the held-out test predictions, so that all three split types are compared under one consistent statistical framework.


\section{Results}
\label{sec:results}

Unless stated otherwise, all numbers below are for the \textbf{active, not toxic} target at $q<0.05$, with the linear probe. Reported $95\,\%$ confidence intervals come from bootstrap-resampling the held-out test predictions for BBBC036v1's official split, and from the 25 samples of the $5\times5$ repeated grouped cross-validation for every chemical- and batch-generalization split (\Cref{sec:metrics}).

\begin{figure}[t]
    \centering
    \includegraphics[width=0.5\linewidth]{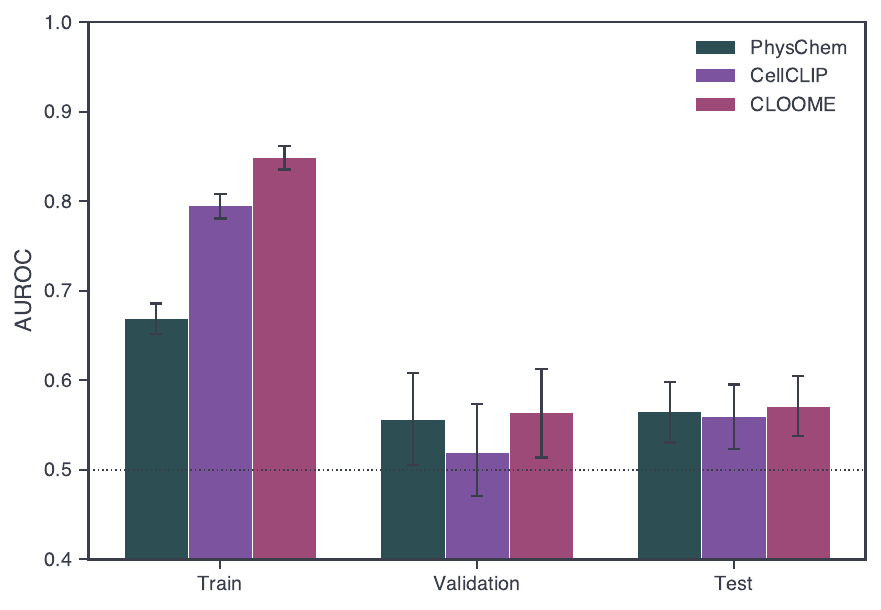}
    \caption{%
        Train, validation, and test AUROC for CLOOME, CellCLIP, and PhysChem on BBBC036v1's official split, target active, not toxic. Bars: mean AUROC across bootstrap resamples of the corresponding fold's predictions; error bars: 95\% CI ($n=1{,}000$ resamples).
    }
    \label{fig:leakage}
\end{figure}

\paragraph{} \textbf{CLOOME and CellCLIP overfit their own pretraining-respecting split.}
\Cref{fig:leakage} shows train, validation, and test AUROC on BBBC036v1's official split. CLOOME drops from 0.849~[0.835, 0.862] on train to 0.571~[0.538, 0.605] on test, a gap of 0.278. CellCLIP drops from 0.795~[0.781, 0.808] to 0.560~[0.524, 0.595], a gap of 0.235. Validation AUROC is far below train AUROC for both, so this is not just test-fold noise. PhysChem's gap is much smaller: 0.669~[0.652, 0.686] on train to 0.565~[0.531, 0.598] on test, a difference of only 0.104. The same pattern shows up on BBBC036v1's chemical and batch cross-validation splits. There, CLOOME scores 0.707 and 0.694, well above its official split test AUROC of 0.571. CellCLIP scores 0.603 and 0.609, above its own 0.560. PhysChem does not show this inflation: 0.552 and 0.552, both below its 0.565. The official split is the only one that guarantees no test compound was part of CLOOME's or CellCLIP's own pretraining set, and it is the split where both encoders do worst. This difference is not explained by the official test set being harder (\Cref{tab:splits}): its test compounds are as close to their training set as in the batch split (median nearest-neighbour Tanimoto similarity 0.49 against 0.51, with 16--17\,\% of test compounds having a close analogue in training against 4\,\% for the chemical split), and label prevalence and median LogP and molecular weight are nearly identical across the three splits.

\begin{figure}[t]
    \centering
    \includegraphics[width=\linewidth]{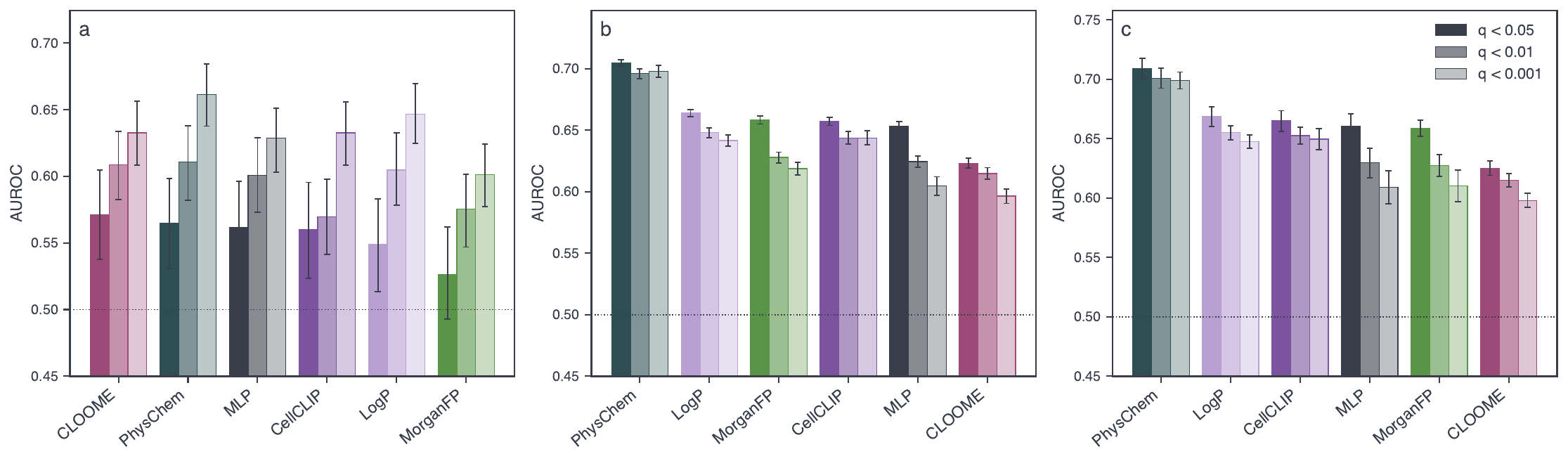}
    \caption{%
        Representation ranking, target active, not toxic. (a)~BBBC036v1 official split; CLOOME's and CellCLIP's pretraining included this split's training compounds (\Cref{fig:leakage}). (b)~JUMP-CP chemical-generalization split. (c)~JUMP-CP batch-generalization split. Representations are sorted by their score at $q<0.05$. For each representation, the three bars, from darkest to lightest, use activity labels at $q<0.05$, $q<0.01$, and $q<0.001$, with toxicity kept at $q<0.05$. Bars: mean AUROC; error bars: 95\% CI (bootstrap resampling in (a), $5\times5$ repeated grouped cross-validation in (b)--(c)).
    }
    \label{fig:ranking}
\end{figure}

\paragraph{} \textbf{PhysChem matches or beats both pretrained encoders, and a non-pretrained control does better than CLOOME itself.} \Cref{fig:ranking} ranks all six representations on BBBC036v1's official split (panel a) and JUMP-CP's chemical and batch splits (panels b-c). On BBBC036v1 (a), all six have heavily overlapping CIs, from CLOOME (0.571~[0.538, 0.605]) and PhysChem (0.565~[0.531, 0.598]) through MLP (0.562~[0.526, 0.596]), CellCLIP (0.560~[0.524, 0.595]), and LogP (0.549~[0.513, 0.583]) down to MorganFP (0.527~[0.493, 0.562]); CLOOME and CellCLIP were pretrained on this split's training compounds (\Cref{fig:leakage}). On JUMP-CP's chemical split (b) (the main population defined in \Cref{sec:data}, excluding the bioactive-library subset), PhysChem is clearly best (0.705~[0.702, 0.707]), followed by LogP, MorganFP, CellCLIP, and MLP (0.664 down to 0.653). CLOOME is last (0.623~[0.619, 0.627]), below the non-pretrained MLP control, with no overlap in CIs. On the batch split (c), CIs widen somewhat but the pattern holds and sharpens: PhysChem is again best (0.709~[0.700, 0.718]), LogP and CellCLIP follow (0.668 and 0.665), and CLOOME (0.625~[0.619, 0.631]) is last, now with no overlap with any other representation, including MorganFP (0.659~[0.652, 0.666]). CellCLIP's ranking varies by split, while CLOOME stays at or near the bottom on both out-of-distribution evaluations, and is significantly worse than every other representation specifically on the batch split. The lighter bars show the same comparison with stricter activity labels, where compounds no longer active become negatives. On BBBC036v1, every representation improves, PhysChem going from 0.565 to 0.661~[0.638, 0.684] at $q<0.001$, although CIs still overlap. On JUMP-CP the ranking holds: PhysChem stays best (0.698~[0.693, 0.703] on the chemical split at $q<0.001$) and CLOOME last (0.596~[0.591, 0.602]). BBBC036v1's activity label is thus permissive, with many weak actives (\appfig{fig:activity}, discussed in \Cref{app:activity}), so we treat JUMP-CP as our main evidence.

\begin{figure}[t]
    \centering
    \includegraphics[width=0.7\linewidth]{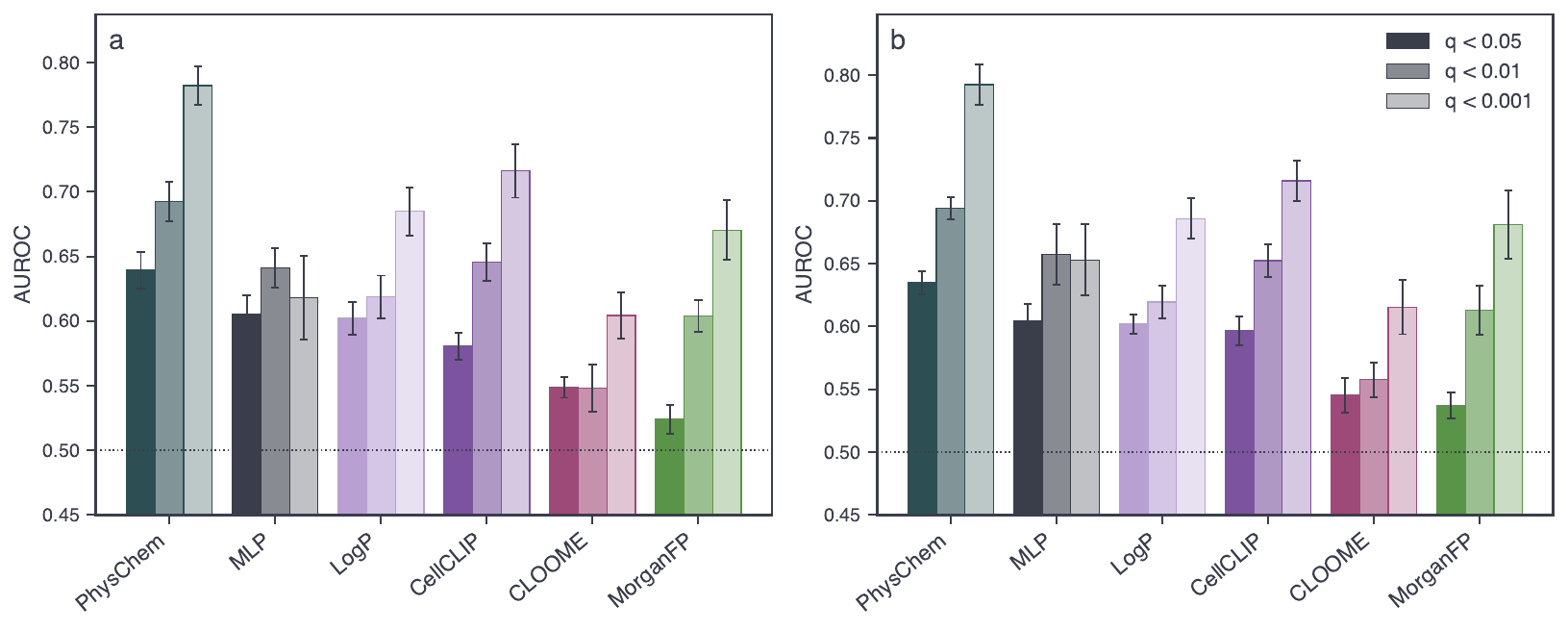}
    \caption{%
        Representation ranking on JUMP-CP's bioactive-library sub-dataset (\textit{source\_7}),
        target active, not toxic. (a)~chemical-generalization split. (b)~plate-generalization split.
        Three bars per representation as in \Cref{fig:ranking}. Bars: mean AUROC; error bars: 95\% CI ($5\times5$ repeated grouped cross-validation).
    }
    \label{fig:bioactive}
\end{figure}

\paragraph{} \textbf{The pattern replicates on a third population.} \Cref{fig:bioactive} repeats the representation-ranking comparison on JUMP-CP's bioactive-library sub-dataset (\textit{source\_7}, \Cref{sec:data}), a curated collection of already-characterized bioactive compounds screened at a different concentration than the rest of JUMP-CP, and so structurally and experimentally distinct from both BBBC036v1 and the rest of JUMP-CP. Its batch-generalization analog uses plate rather than source as the grouping variable, since this subset spans only one source (\Cref{sec:metrics}). PhysChem is again clearly best on both its chemical (0.639~[0.625, 0.653]) and plate (0.635~[0.625, 0.644]) splits, and CLOOME is again at the bottom (0.549~[0.541, 0.556] and 0.545~[0.531, 0.559]), together with MorganFP (0.524 and 0.537). With stricter activity labels, PhysChem's lead grows (0.782~[0.767, 0.797] on the chemical split at $q<0.001$) and CLOOME stays last. The same ranking holding on a different population from either main dataset is evidence that it reflects something about these representations, not an artifact of one screen's chemistry or experimental design.

\paragraph{} \textbf{A non-linear probe does not change the comparison.} With the MLP probe instead of the linear one (\appfig{fig:probes}), CLOOME, CellCLIP, and PhysChem all stay between 0.56 and 0.57 on BBBC036v1's official split, with overlapping CIs. On JUMP-CP's chemical split, the MLP probe changes little: PhysChem reaches 0.716~[0.712, 0.720], against 0.654 for CellCLIP and 0.627 for CLOOME. The largest gains are on \textit{source\_7}, where CLOOME goes from 0.549 to 0.607 and CellCLIP from 0.581 to 0.610, but PhysChem stays ahead (0.643~[0.629, 0.656]), and CLOOME with an MLP probe only reaches the level of the MLP control trained from scratch on its input fingerprint (0.606).

\begin{figure}[t]
    \centering
    \includegraphics[width=\linewidth]{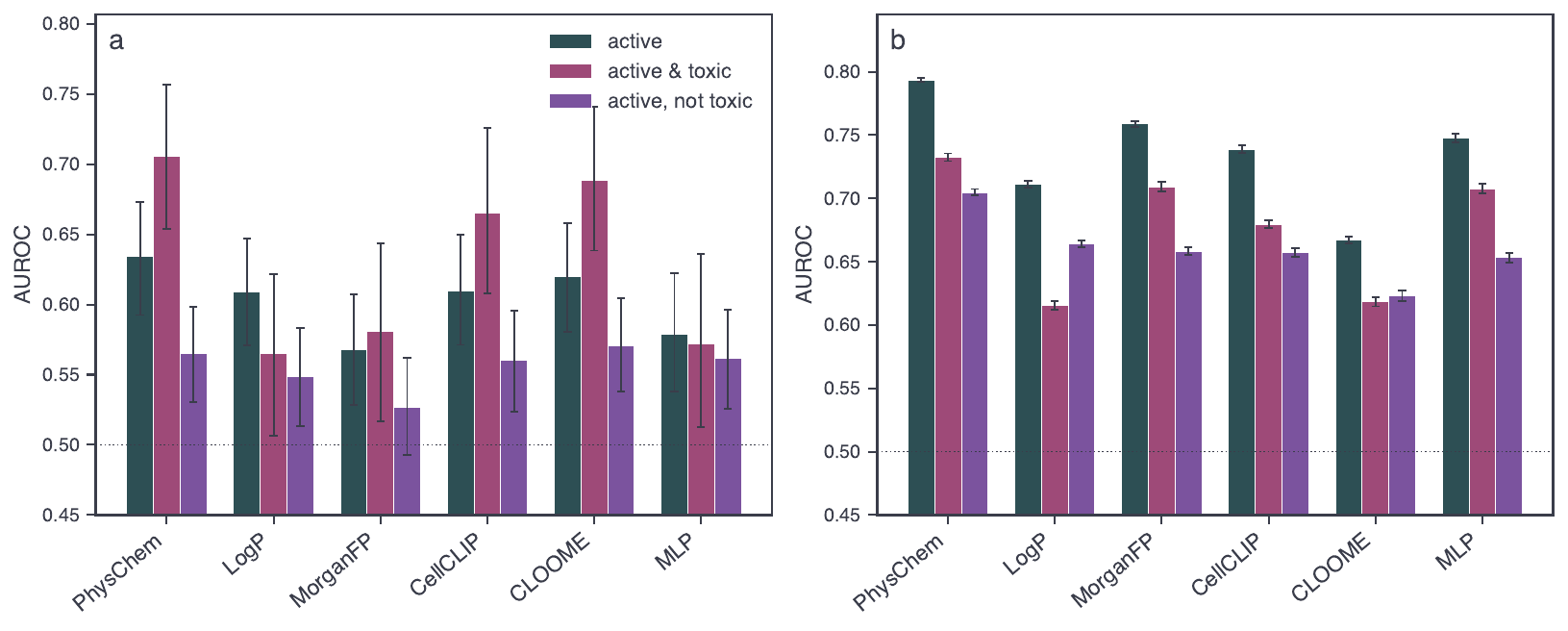}
    \caption{%
        Mean test AUROC by prediction target (active; active and toxic; active, not toxic) for all
        six representations, on (a)~BBBC036v1's official split and (b)~JUMP-CP's chemical-generalization split. Error bars: 95\% CI, as in \Cref{fig:ranking}.
    }
    \label{fig:task-difficulty}
\end{figure}

\paragraph{} \textbf{Toxicity is easier to predict than true phenotypic activity using almost every representation.} \Cref{fig:task-difficulty} compares all six representations across all three targets. On BBBC036v1's official split, active-not-toxic is the hardest target for all six representations. PhysChem, for example, reaches 0.706 on active-and-toxic and 0.634 on plain active, but only 0.565 on active-not-toxic. On JUMP-CP's chemical split, the same order holds for four of six representations (PhysChem, MorganFP, CellCLIP, MLP). For LogP and CLOOME, active-and-toxic is instead the hardest, or tied with active-not-toxic (CLOOME: 0.618 and 0.623). Plain activity, which is not controlled for toxicity, is the easiest target for every representation on JUMP-CP. On BBBC036v1, where only $6.3\,\%$ of actives are toxic, active-and-toxic is instead the easiest target for four of six representations (PhysChem, MorganFP, CLOOME, CellCLIP). This pattern holds as strongly for the structure-only baselines as for CLOOME and CellCLIP, so it reflects the activity/toxicity confound itself, not something specific to phenotype-pretrained encoders.

\paragraph{} \textbf{PhysChem's signal is mostly lipophilicity.} Lipophilicity is a classic correlate of nonspecific cytotoxicity \cite{lu2012contribution}, and it also carries most of PhysChem's signal (\appfig{fig:descriptors}). On JUMP-CP, LogP alone reaches 0.664~[0.661, 0.667], against 0.705 for all 217 descriptors, and adding size, polarity, and basic amines only brings it to 0.675. On BBBC036v1, all descriptor sets give similar AUROCs with overlapping CIs.


\section{Discussion}
\label{sec:discussion}

Our results highlight the challenges of contrastive molecule-morphology pretraining. For CLOOME and CellCLIP, the apparent skill largely reflects being tested on compounds close to their own pretraining data (\Cref{fig:leakage}); once that is controlled for, neither encoder adds anything over physicochemical
descriptors that use no phenotypic data. These conclusions concern these two encoders under our protocol; they do not establish a general limitation of phenotype-informed pretraining, and the same protocol can be applied directly to future encoders.

The MLP control suggests why this is hard rather than incidental: a network pretrained contrastively on one screen's molecule-morphology pairs is fit to that screen's comparatively narrow slice of chemical space, and nothing about the contrastive objective guarantees the resulting mapping holds once the input distribution shifts; on JUMP-CP, a non-pretrained control matching CLOOME's input and layer count generalizes better than CLOOME's pretrained version does, and the same holds on JUMP-CP's structurally distinct bioactive-library subset (\Cref{fig:ranking,fig:bioactive}). This is not a consequence of the linear probe: with an MLP on top of its frozen embedding, CLOOME only matches the control trained from scratch (\appfig{fig:probes}). This echoes \citet{rottach2026moljepa}, who find that including CLOOME as an input modality in a multimodal pretraining framework can decrease downstream performance, and \citet{fradkin2024molecules}'s own comparison against CLOOME in the paper introducing MolPhenix, which finds it trails their improved model by $8.78\times$ in top-1\% recall on an unseen dataset. CellCLIP does not fail in quite the same way, which suggests the specific pretraining scheme and architecture matter and that not every contrastive encoder collapses out-of-distribution to the same degree.

PhysChem's advantage, being mostly lipophilicity (\appfig{fig:descriptors}), likely reflects general cellular perturbation rather than specific mechanisms, and the pretrained encoders do not exceed even this simple signal. It also suggests a possible limitation of fingerprint-based encoders: substructure fingerprints do not encode bulk properties such as lipophilicity directly, which may partly explain why the three representations built on Morgan fingerprints (MorganFP, the MLP control, and CLOOME) trail PhysChem.

Toxicity is generally easier to predict than genuine, specific phenotypic activity (\Cref{fig:task-difficulty}), and held-out performance can just as easily reflect batch recognition as biology, so a benchmark that does not explicitly control for both risks rewarding models for the wrong reasons, independent of whether they involve any phenotype-derived pretraining at all. Confound- and leakage-aware evaluation needs to become a basic reporting standard before phenotypically-informed models can be trusted as surrogates for cellular response, rather than for toxicity or dataset provenance.

These results also motivate a hypothesis for phenotype-driven molecular generation, which we did not test: a generative loop that optimizes against an encoder output will exploit whatever the encoder actually rewards, and if that is structural similarity to a pretraining set or toxicity rather than an interesting phenotype, generation could drift there instead. The same concern applies in the other direction, generating morphology from structure: since most perturbations in a screen produce little or no phenotypic change, a model such as MorphoDiff \cite{navidi2025morphodiff} can score well in aggregate without ever being shown to capture actual structure-activity relationships for the active minority, unless activity itself is accounted for during evaluation.

\subsection{Limitations and future work}
\label{sec:limitations}

Several limitations bound how far these results generalize. CLOOME and CellCLIP are both pretrained on BBBC036v1, so we cannot separate a general property of contrastive phenotype-pretraining from a property of this one, narrow pretraining set; larger or more chemically diverse pretraining corpora might not show the same collapse. BBBC036v1's official test fold is small (2,131 compounds, \Cref{tab:splits}), which is why several representations remain statistically indistinguishable. JUMP-CP's batch split rests on a modest number of unevenly sized sources even after our mode-source grouping, which is why its confidence intervals stay wider than the chemical split's, and these sources differ in laboratory, compounds, and conditions at the same time; source-wise statistics and a leave-one-source-out analysis would make the cross-laboratory result more transparent. Every representation is scored through a linear probe or, for the MLP control and the MLP probe, a shallow network; neither changes the ranking, but direct fine-tuning of CLOOME and CellCLIP on these labels could still narrow the gaps. Labels are also not derived identically across datasets: JUMP-CP's activity labels come from the community-published \textit{jump\_rr} $p$-values, not our own pipeline, so some cross-dataset differences could reflect this rather than biology. Toxicity is estimated here with a cell-count-based proxy (i.e., a permutation test on cell-count deviation from negative control) and is not an independent cytotoxicity measurement: a lower cell count can also reflect cytostasis or reduced proliferation, and some toxic responses may leave cell count unchanged.

Two things remain untested rather than shown. First, the molecular-generation argument in \Cref{sec:discussion} is a hypothesis: we did not run a generative model against these encoders as a reward signal, so whether generation actually drifts toward familiar or toxic compounds remains to be checked directly. Second, concentration. Both activity and toxicity are dose-dependent, and JUMP-CP's bioactive-library subset hints at this: screened at $0.625\,\mu$M, it shows a markedly different activity and toxicity profile from the rest of JUMP-CP's diversity sources, screened at $10\,\mu$M. But neither dataset here varies concentration for the same compound, so we cannot separate concentration's own contribution from the compound-population differences that come with it. A dataset that does, such as RxRx3-core \cite{kraus2025rxrx3} (8 concentrations spanning 0.0025--10~$\mu$M), would let this be tested directly.


\section{Conclusion}
\label{sec:conclusion}

We evaluated whether two phenotype-pretrained molecular encoders, CLOOME and CellCLIP, can predict phenotypic activity from structure alone, using an evaluation protocol designed to rule out two specific confounds: pretraining-boundary leakage and the correlation between activity and toxicity. Under that protocol, both encoders lose their apparent advantage. Neither outperforms plain physicochemical descriptors on compounds outside their own pretraining set, also with stricter activity labels or a non-linear probe, and CLOOME performs worse than a non-pretrained control matching its input and layer count on an independent screen. Toxicity, meanwhile, is easier to predict than true phenotypic activity for almost every representation tested, pretrained or not, which means activity-prediction results that do not separate the two should be read with caution regardless of the encoder involved. These findings do not rule out phenotype-pretrained encoders as useful tools, but they do mean the two tested here should not yet be trusted as phenotypic surrogates, and that claims of this kind, for these or future encoders, should be checked under a comparably leakage-aware, confound-controlled evaluation.


\section*{Author contributions}
\textbf{TC:} Conceptualization, Methodology, Software, Formal analysis, Investigation, Visualization,
Writing --- original draft.
\textbf{RM:} Conceptualization, Supervision, Funding acquisition, Resources, Writing --- review \& editing.


\begin{ack}
TC and RM acknowledge funding provided by the Wallenberg AI, Autonomous Systems, and Software Program (WASP), supported by the Knut and Alice Wallenberg Foundation. Computational resources were provided by the National Academic Infrastructure for Supercomputing in Sweden (NAISS), funded by the Swedish Research Council. Computational resources were provided on the Berzelius system funded by the Knut and Alice Wallenberg foundation and operated by NAISS.
\end{ack}


\section*{Declaration of competing interest}
The authors declare no competing interests.


\section*{Software and data availability}
Code is available at \url{https://github.com/cctelio/predicting-phenotypic-activity}. The full analysis, from obtaining raw data to reproducing every figure and table in this paper, reruns in a few hours on a standard workstation CPU.


\section*{Declaration of Generative AI and AI-assisted technologies in the writing process}
During the preparation of this work, the authors used Claude (Anthropic) to help draft and edit
the manuscript text, generate plotting code for figures made from the study's own data, and assist
with coding throughout the analysis. The authors reviewed and edited all resulting content and
take full responsibility for the content of this article.

\bibliographystyle{unsrtnat}
\bibliography{references}

@article{seal2026counting,
  title={Counting cells can accurately predict small-molecule bioactivity benchmarks},
  author={Seal, Srijit and Dee, William and Shah, Adit and Cerisier, Natacha and Zhang, Andrew and Miglietta, Esteban and Titterton, Katherine and Cabrera, {\'A}ngel Alexander and Boiko, Daniil and Beatson, Alex and others},
  journal={Nature Communications},
  volume={17},
  number={1},
  pages={2436},
  year={2026},
  publisher={Nature Publishing Group UK London}
}

@article{seal2025cell,
  title={{Cell Painting}: a decade of discovery and innovation in cellular imaging},
  author={Seal, Srijit and Trapotsi, Maria-Anna and Spjuth, Ola and Singh, Shantanu and Carreras-Puigvert, Jordi and Greene, Nigel and Bender, Andreas and Carpenter, Anne E},
  journal={Nature Methods},
  volume={22},
  number={2},
  pages={254--268},
  year={2025},
  publisher={Nature Publishing Group US New York}
}

@article{fredin2024cell,
  title={{Cell Painting}-based bioactivity prediction boosts high-throughput screening hit-rates and compound diversity},
  author={Fredin Haslum, Johan and Lardeau, Charles-Hugues and Karlsson, Johan and Turkki, Riku and Leuchowius, Karl-Johan and Smith, Kevin and M{\"u}llers, Erik},
  journal={Nature Communications},
  volume={15},
  number={1},
  pages={3470},
  year={2024},
  publisher={Nature Publishing Group UK London}
}

@misc{rottach2026moljepa,
      title={{Mol-JEPA}: A multimodal Joint Embedding Predictive Architecture for Molecules}, 
      author={Florian Rottach and Sebastian Schieferdecker and William Rudman and Randall Balestriero and Carsten Eickhoff},
      year={2026},
      eprint={2608.22642},
      archivePrefix={arXiv},
      primaryClass={cs.LG},
      url={https://arxiv.org/abs/2608.22642}, 
}

@article{moshkov2023predicting,
  title={Predicting compound activity from phenotypic profiles and chemical structures},
  author={Moshkov, Nikita and Becker, Tim and Yang, Kevin and Horvath, Peter and Dancik, Vlado and Wagner, Bridget K and Clemons, Paul A and Singh, Shantanu and Carpenter, Anne E and Caicedo, Juan C},
  journal={Nature Communications},
  volume={14},
  number={1},
  pages={1967},
  year={2023},
  publisher={Nature Publishing Group UK London}
}

@article{marin2023cell,
  title={Cell morphology-guided de novo hit design by conditioning {GANs} on phenotypic image features},
  author={Marin Zapata, Paula A and M{\'e}ndez-Lucio, Oscar and Le, Tuan and Beese, Carsten J{\"o}rn and Wichard, J{\"o}rg and Rouqui{\'e}, David and Clevert, Djork-Arn{\'e}},
  journal={Digital Discovery},
  volume={2},
  number={1},
  pages={91--102},
  year={2023},
  publisher={The Royal Society of Chemistry}
}

@article{sanchez2023cloome,
  title={{CLOOME}: contrastive learning unlocks bioimaging databases for queries with chemical structures},
  author={Sanchez-Fernandez, Ana and Rumetshofer, Elisabeth and Hochreiter, Sepp and Klambauer, G{\"u}nter},
  journal={Nature Communications},
  volume={14},
  number={1},
  pages={7339},
  year={2023},
  publisher={Nature Publishing Group UK London}
}

@article{lu2026cellclip,
  title={{CellCLIP}-Learning Perturbation Effects in Cell Painting via Text-Guided Contrastive Learning},
  author={Lu, Mingyu and Weinberger, Ethan and Kim, Chanwoo and Lee, Su-In},
  journal={Advances in Neural Information Processing Systems},
  volume={38},
  pages={124505--124537},
  year={2025}
}

@article{lu2012contribution,
  title={The contribution of physicochemical properties to multiple in vitro cytotoxicity endpoints},
  author={Lu, Shuyan and Jessen, Bart and Strock, Christopher and Will, Yvonne},
  journal={Toxicology in Vitro},
  volume={26},
  number={4},
  pages={613--620},
  year={2012},
  publisher={Elsevier}
}

@article{moshkov2024learning,
  title={Learning representations for image-based profiling of perturbations},
  author={Moshkov, Nikita and Bornholdt, Michael and Benoit, Santiago and Smith, Matthew and McQuin, Claire and Goodman, Allen and Senft, Rebecca A and Han, Yu and Babadi, Mehrtash and Horvath, Peter and others},
  journal={Nature Communications},
  volume={15},
  number={1},
  pages={1594},
  year={2024},
  publisher={Nature Publishing Group UK London}
}

@article{kalinin2025versatile,
  title={A versatile information retrieval framework for evaluating profile strength and similarity},
  author={Kalinin, Alexandr A and Arevalo, John and Serrano, Erik and Vulliard, Loan and Tsang, Hillary and Bornholdt, Michael and Mu{\~n}oz, Al{\'a}n F and Sivagurunathan, Suganya and Rajwa, Bartek and Carpenter, Anne E and others},
  journal={Nature Communications},
  volume={16},
  number={1},
  pages={5181},
  year={2025},
  publisher={Nature Publishing Group UK London}
}

@article{bray2017dataset,
  title={A dataset of images and morphological profiles of 30,000 small-molecule treatments using the Cell Painting assay},
  author={Bray, Mark-Anthony and Gustafsdottir, Sigrun M and Rohban, Mohammad H and Singh, Shantanu and Ljosa, Vebjorn and Sokolnicki, Katherine L and Bittker, Joshua A and Bodycombe, Nicole E and Dan{\v{c}}{\'\i}k, Vlado and Hasaka, Thomas P and others},
  journal={Gigascience},
  volume={6},
  number={12},
  pages={giw014},
  year={2017},
  publisher={Oxford University Press}
}

@article{ando2017improving,
  title={Improving phenotypic measurements in high-content imaging screens},
  author={Ando, D Michael and McLean, Cory Y and Berndl, Marc},
  journal={BioRxiv},
  pages={161422},
  year={2017},
  publisher={Cold Spring Harbor Laboratory}
}

@article{chandrasekaran2023jump,
  title={{JUMP} {Cell Painting} dataset: morphological impact of 136,000 chemical and genetic perturbations},
  author={Chandrasekaran, Srinivas Niranj and Ackerman, Jeanelle and Alix, Eric and Ando, D Michael and Arevalo, John and Bennion, Melissa and Boisseau, Nicolas and Borowa, Adriana and Boyd, Justin D and Brino, Laurent and others},
  journal={BioRxiv},
  pages={2023--03},
  year={2023},
  publisher={Cold Spring Harbor Laboratory}
}

@article{wildman1999prediction,
  title={Prediction of physicochemical parameters by atomic contributions},
  author={Wildman, Scott A and Crippen, Gordon M},
  journal={Journal of Chemical Information and Computer Sciences},
  volume={39},
  number={5},
  pages={868--873},
  year={1999},
  publisher={ACS Publications}
}

@article{butina1999unsupervised,
  title={Unsupervised data base clustering based on daylight's fingerprint and Tanimoto similarity: A fast and automated way to cluster small and large data sets},
  author={Butina, Darko},
  journal={Journal of Chemical Information and Computer Sciences},
  volume={39},
  number={4},
  pages={747--750},
  year={1999},
  publisher={ACS Publications}
}

@article{ash2025practically,
  title={Practically significant method comparison protocols for machine learning in small molecule drug discovery},
  author={Ash, Jeremy R and Wognum, Cas and Rodr{\'\i}guez-P{\'e}rez, Raquel and Aldeghi, Matteo and Cheng, Alan C and Clevert, Djork-Arn{\'e} and Engkvist, Ola and Fang, Cheng and Price, Daniel J and Hughes-Oliver, Jacqueline M and others},
  journal={Journal of Chemical Information and Modeling},
  volume={65},
  number={18},
  pages={9398--9411},
  year={2025},
  publisher={ACS Publications}
}

@article{munoz2026jump,
  title={{JUMP-lite}: Compact, reproducible benchmarking of cell representations},
  author={Mu{\~n}oz, Al{\'a}n F and Haslum, Johan Fredin and Shen, Runxi and Carpenter, Anne E and Singh, Shantanu},
  journal={arXiv preprint arXiv:2608.07632},
  year={2026}
}

@article{kraus2025rxrx3,
  title={{RxRx3-core}: benchmarking drug-target interactions in high-content microscopy},
  author={Kraus, Oren and Comitani, Federico and Urbanik, John and Kenyon-Dean, Kian and Arumugam, Lakshmanan and Saberian, Saber and Wognum, Cas and Celik, Safiye and Haque, Imran S},
  journal={arXiv preprint arXiv:2503.20158},
  year={2025}
}

@article{fradkin2024molecules,
  title={How molecules impact cells: Unlocking contrastive phenomolecular retrieval},
  author={Fradkin, Philip and Azadi, Puria and Suri, Karush and Wenkel, Frederik and Bashashati, Ali and Sypetkowski, Maciej and Beaini, Dominique},
  journal={Advances in Neural Information Processing Systems},
  volume={37},
  pages={110667--110701},
  year={2024}
}

@article{harrison2026biologically,
  title={Biologically-driven generative chemistry: using biological data to guide {\sl de novo} drug design},
  author={Harrison, Philip John and Mercado, Roc{\'\i}o},
  year={2026},
  journal={ChemRxiv},
}

@inproceedings{navidi2025morphodiff,
  title={{MorphoDiff}: Cellular morphology painting with diffusion models},
  author={Navidi, Zeinab and Ma, Jun and Miglietta, Esteban and Liu, Le and Carpenter, Anne and Cimini, Beth and Haibe-Kains, Benjamin and Wang, Bo},
  booktitle={International Conference on Learning Representations},
  volume={2025},
  pages={11613--11637},
  year={2025}
}


\beginappendix
\crefalias{section}{appendix}
\crefalias{subsection}{appendix}
\setcounter{figure}{0}
\setcounter{table}{0}
\renewcommand{\thefigure}{A\arabic{figure}}
\renewcommand{\thetable}{A\arabic{table}}

\section{Model details}
\label{app:model-details}

\Cref{tab:representations} records how each representation, initially introduced in \Cref{sec:model}, is used to make predictions. Five of the six representations are frozen and read out by an $\ell_2$-regularized logistic regression, so the probe is the only thing fitted on our labels. The MLP control is not: it is trained end to end on the Morgan fingerprint, without morphology pretraining.
 
The representations span three orders of magnitude in dimensionality, from a single Crippen logP value to a 2{,}048-bit fingerprint, so the $\ell_2$ penalty is selected per fold and per representation on an inner training/validation split over a 3-point grid, $C \in \{0.01, 1, 100\}$. Features are standardized (zero mean, unit variance, fit on the training fold only); the logistic regression probe uses \texttt{solver="lbfgs"}, \texttt{max\_iter=1000}, and \texttt{class\_weight="balanced"}. Contrastive embeddings are taken from each encoder's own final 512-d output (CLOOME's last MLP layer; CellCLIP's linear projection of BERT's pooled output), not an earlier intermediate representation, and are L2-normalized, matching each model's own contrastive training objective.

\begin{table}[h]
\centering
\footnotesize
\setlength{\tabcolsep}{4pt}
\caption{Summary of every representation and model evaluated in this work for the various sets of labels.
Five representations are fixed and read out by an $\ell_2$-regularized logistic regression probe, the only thing fitted on our labels. The MLP control is the exception: it is trained directly on our labels from CLOOME's Morgan fingerprint input, without morphology pretraining.}
\label{tab:representations}
\begin{tabular}{@{}l >{\raggedright\arraybackslash}p{0.175\linewidth} c
                >{\raggedright\arraybackslash}p{0.185\linewidth}
                >{\raggedright\arraybackslash}p{0.175\linewidth}@{}}
\toprule
\textbf{Representation tested} & \textbf{Input} & \textbf{Dim.} &
\textbf{Source} & \textbf{Model} \\
\midrule
PhysChem   & SMILES & 217 & RDKit descriptors, deterministic & Logistic regression \\
\addlinespace[2pt]
LogP       & SMILES & 1 & Crippen logP, deterministic & Logistic regression \\
\addlinespace[2pt]
MorganFP   & SMILES & 2{,}048 & Morgan fingerprint, $r=2$ & Logistic regression \\
\addlinespace[2pt]
CLOOME     & Morgan FP, $r=3$, 1{,}024 bits & 512 & Frozen pretrained checkpoint & Logistic regression \\
\addlinespace[2pt]
CellCLIP   & Templated SMILES prompt & 512 & Frozen pretrained checkpoint & Logistic regression \\
\addlinespace[2pt]
MLP        & Morgan FP, $r=3$, 1{,}024 bits & --- & --- & Entire network (end-to-end) \\
\bottomrule
\end{tabular}
\end{table}

The MLP control takes the same chiral Morgan fingerprint input as CLOOME (radius 3, 1{,}024 bits) and has the same number of fully connected layers (four), but receives no contrastive pretraining. Its layers taper (hidden widths 512, 256, 128, and 64, ReLU activations), unlike CLOOME's own published configuration, which uses a constant hidden dimension of 1{,}024 throughout (per its released model config, \texttt{RN50.json}). The final layer is a single logit, produced end-to-end. CLOOME itself has no classifier: it is a contrastive dual-encoder model, so its 512-d embedding is instead read out with the same logistic-regression probe used for every other frozen representation (\Cref{sec:model}). Weights use PyTorch's default \texttt{nn.Linear} initialization; no custom initialization scheme is applied. The MLP probe uses the same network and training; only its input layer differs, taking the frozen CLOOME (512-d), CellCLIP (512-d), or PhysChem (217-d) representation instead of the 1{,}024-bit fingerprint. In the descriptor analysis, basic amines are counted with a SMARTS pattern, since they are not one of the RDKit descriptors.

Training uses AdamW at an initial learning rate of $10^{-3}$, held constant, batch size 256, weight decay $10^{-4}$, and dropout 0.4 applied after each hidden layer's activation. The objective is binary cross-entropy with logits, with a per-fold class weighting (\texttt{pos\_weight} set to the inverse class-frequency ratio of the training fold). Training runs for at most 150 epochs with early stopping on validation AUROC evaluated on the inner validation split, patience 15 epochs, and the checkpoint with the best validation AUROC is used for the held-out fold.

The inner validation split holds out $20\,\%$ of each training fold (stratified by label) and is drawn at random, not with the same grouping constraint as the outer split, so a structurally similar compound can occasionally inform early stopping even though it cannot inform the held-out fold's test predictions; this is a real, if comparatively mild, gap relative to the outer split's grouping guarantee. Hyperparameters (dropout, weight decay, learning rate, batch size) are fixed a priori rather than searched per fold or per representation. Every fold is trained with a single random seed, drawn from the corresponding cross-validation repeat; statistical replication comes from the $5\times5$ repeated grouped cross-validation itself (\Cref{sec:metrics}), not from separate seed-averaging within a fold. Training is not bit-reproducible on CPU, and reruns move the MLP's AUROC by up to about 0.02. For the stricter activity thresholds, BBBC036v1's activity $p$-values are recomputed with $10^6$ permutations, which leaves the fraction active at $q<0.05$ unchanged ($91.3$--$91.4\,\%$); JUMP-CP's published $p$-values already resolve $q<0.001$. Software versions: \texttt{rdkit}~2025.03, \texttt{pytorch}~2.8, \texttt{scikit-learn}~1.7 (see \texttt{environment.yml} for the full, pinned dependency list).

\section{Additional results}
\label{app:additional-results}

\subsection{Activity in BBBC036v1 and JUMP-CP}
\label{app:activity}

\begin{figure}[H]
    \centering
    \includegraphics[width=0.84\linewidth]{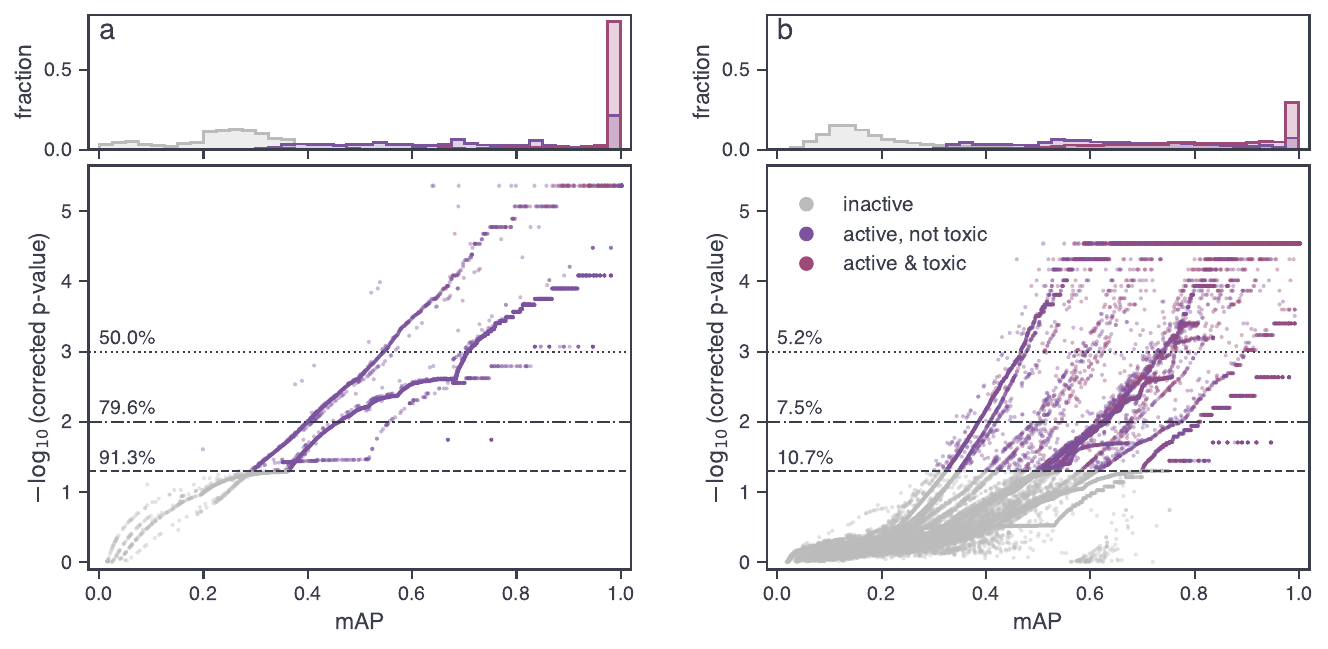}
    \caption{%
        Phenotypic activity of (a)~BBBC036v1 and (b)~JUMP-CP. Bottom: mAP against $-\log_{10}$ of the Benjamini-Hochberg-corrected $p$-value, one point per compound; inactive compounds ($q\geq0.05$) in grey, active compounds by toxicity call. Horizontal lines mark $q=0.05$ (dashed), $0.01$ (dash-dot), and $0.001$ (dotted), each labeled with the fraction of compounds called active at that threshold. The discrete traces follow the permutation test's pair configurations, discussed in \Cref{app:activity}. Top: mAP distribution of each class, normalized within the class. BBBC036v1 $p$-values use $10^6$ permutations. JUMP-CP shows all compounds with a published \textit{jump\_rr} $p$-value after removing the overlap with BBBC036v1, before class balancing and including \textit{source\_7}.
    }
    \label{fig:activity}
\end{figure}

\Cref{fig:activity} plots each compound's mAP against the corrected $p$-value of its permutation test, the two quantities the activity label of \Cref{sec:data} is built from. Almost all of BBBC036v1 sits above the $q<0.05$ line, most of it at a high mAP, while the bulk of JUMP-CP falls below it: a curated repurposing collection against a chemically broad discovery screen, and a single-laboratory screen whose replicates agree closely against one whose replicates are spread over several sources. The mAP values are not comparable between the two, since each averages precision over a different number of replicate and control wells.

The scatter also breaks into discrete traces rather than a single curve, which we read as an artefact of the test rather than a property of the biology. \texttt{copairs} derives a separate null for each pair configuration, that is, for the number of replicate and total pairs a compound contributes, and within a configuration the mapping from mAP to $p$-value is monotone, so each configuration would be expected to draw its own curve, shifted right when a compound has fewer replicates and needs a larger effect to reach the same significance. On the same reading, the flat ceiling is where compounds beat all $N$ permutations and share the smallest attainable $p$-value, and the staircases and the spike at $\mathrm{mAP}=1$ follow from the few distinct values average precision can take over a handful of replicate wells. In both screens the toxic actives sit at the high-mAP end, which we take to reflect cell loss producing the largest and most reproducible deviation from control.

\subsection{Linear and non-linear probes}
\label{app:probes}

\begin{figure}[H]
    \centering
    \includegraphics[width=0.8\linewidth]{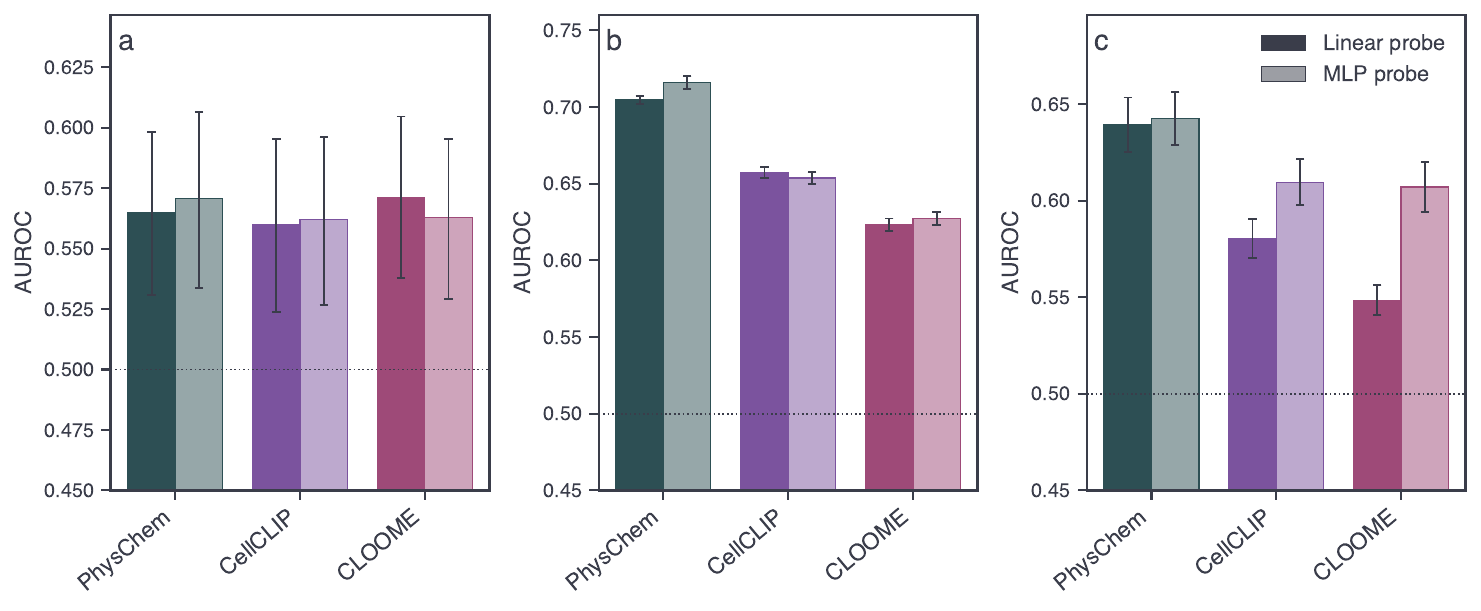}
    \caption{%
        Linear probe (darker) and MLP probe (lighter) on the frozen PhysChem, CellCLIP, and CLOOME representations, target active, not toxic. (a)~BBBC036v1 official split. (b)~JUMP-CP chemical-generalization split. (c)~JUMP-CP bioactive-library (\textit{source\_7}) chemical-generalization split. Bars: mean AUROC; error bars: 95\% CI (bootstrap resampling in (a), $5\times5$ repeated grouped cross-validation in (b)--(c)).
    }
    \label{fig:probes}
\end{figure}

Reading every representation with a linear probe could understate an embedding whose phenotypic information is present but not linearly accessible. \Cref{fig:probes} repeats the comparison for PhysChem, CellCLIP, and CLOOME with the MLP probe of \Cref{app:model-details}, the same network and training as the MLP control with only its input layer resized. On BBBC036v1's official split the three stay between 0.56 and 0.57 with overlapping confidence intervals, as with the linear probe, and on JUMP-CP's chemical split PhysChem stays ahead at 0.716~[0.712, 0.720] against 0.654 for CellCLIP and 0.627 for CLOOME. The clearest gains are on the bioactive-library subset, where CLOOME improves from 0.549 to 0.607 and CellCLIP from 0.581 to 0.610, but PhysChem remains ahead (0.643~[0.629, 0.656]) and CLOOME only reaches the level of the MLP control trained from scratch on the same fingerprint input (0.606).

\subsection{PhysChem descriptor importance}
\label{app:descriptors}

\begin{figure}[H]
    \centering
    \includegraphics[width=0.5\linewidth]{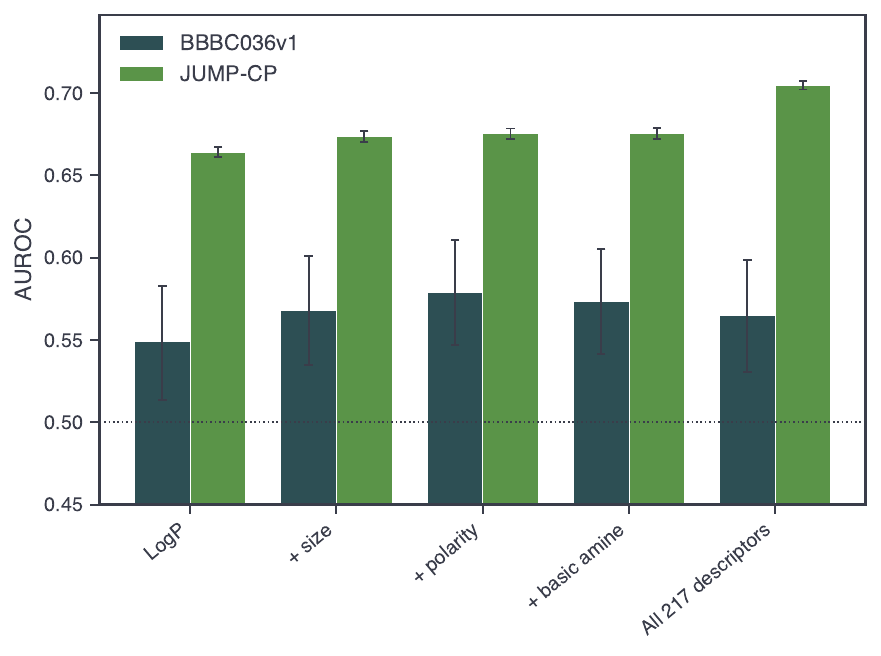}
    \caption{%
        AUROC of the probe refit on a nested set of descriptors and on all 217 RDKit descriptors (PhysChem), target active, not toxic, on BBBC036v1's official split and JUMP-CP's chemical-generalization split. Error bars: 95\% CI, as in \Cref{fig:ranking}.
    }
    \label{fig:descriptors}
\end{figure}

PhysChem is the strongest representation on both JUMP-CP splits, so it is worth asking which of its 217 descriptors that rests on. \Cref{fig:descriptors} refits the same probe on a nested sequence of interpretable descriptors: Crippen logP alone, then with molecular weight and heavy-atom count, then TPSA and hydrogen-bond donors and acceptors, then basic amines. On JUMP-CP's chemical split logP alone already reaches 0.664~[0.661, 0.667] of the 0.705 obtained with all 217, the three further groups add only 0.011, and the remaining 0.030 is spread across the rest of the descriptor set rather than carried by any single interpretable property. On BBBC036v1's official split every nested set gives an indistinguishable AUROC, consistent with that screen's permissive activity label (\Cref{app:activity}) leaving little to separate. Since lipophilicity is a classic correlate of nonspecific cytotoxicity \cite{lu2012contribution}, most of PhysChem's advantage resting on it bears directly on the interpretation in \Cref{sec:discussion}.

\end{document}